\pdfoutput=1

\documentclass[11pt]{article}

\usepackage[final]{acl}

\usepackage{times}
\usepackage{latexsym}

\usepackage{multirow}

\usepackage[T1]{fontenc}

\usepackage[utf8]{inputenc}

\usepackage{microtype}

\usepackage{inconsolata}

\usepackage{graphicx}

\usepackage{longtable}
\usepackage{booktabs}
\usepackage{tcolorbox}

\title{UniToMBench: Integrating Perspective-Taking to Improve Theory of Mind in LLMs}

\usepackage{authblk}
\author{
    \textbf{Prameshwar Thiyagarajan} \quad
    \textbf{Vaishnavi Parimi} \quad
    \textbf{Shamant Sai} \\
    \textbf{Soumil Garg} \quad
    \textbf{Zhangir Meirbek} \quad
    \textbf{Nitin Yarlagadda} \quad
    \textbf{Kevin Zhu} \quad
    \textbf{Chris Kim}
}
\affil{Algoverse AI Research}
\affil{\texttt{kevin@algoverse.us, chris.c.kim@stanford.edu}}
\date{}

\begin{document}
\maketitle
\begin{abstract}
Theory of Mind (ToM), the ability to understand the mental states of oneself and others, remains a challenging area for large language models (LLMs), which often fail to predict human mental states accurately. \noindent In this paper, we introduce \textbf{UniToMBench}, a unified benchmark that integrates the strengths of SimToM \citep{wilf2023thinktwiceperspectivetakingimproves} and TOMBENCH \citep{chen2024tombenchbenchmarkingtheorymind} to systematically improve and assess ToM capabilities in LLMs by integrating multi-interaction task designs and evolving story scenarios. Supported by a custom dataset of over 1,000 hand-written scenarios, UniToMBench combines perspective-taking techniques with diverse evaluation metrics to better stimulate social cognition in LLMs. Through evaluation, we observe that while models like GPT-4o and GPT-4o Mini \cite{openai2024gpt4technicalreport} show consistently high accuracy in tasks involving emotional and belief-related scenarios, with results usually above 80\%, there is significant variability in their performance across knowledge-based tasks. These results highlight both the strengths and limitations of current LLMs in ToM-related tasks, underscoring the value of UniToMBench as a comprehensive tool for future development. Our code is \href{https://github.com/Shamant/unifiedtombenchmark.git}{publicly available}.
\end{abstract}

\section{Introduction}

Large Language Models (LLMs) like GPT-4 \citep{openai2024gpt4technicalreport} and Llama \citep{touvron2023llama2openfoundation} have made significant advancements in tasks like question answering \citep{article} and supporting industries through applications such as customer support, translation, and research assistance \citep{10.36227/techrxiv.23589741.v6}. However, LLMs face challenges in domain-specific tasks, often hallucinating or struggling with ethical issues like bias and lack of explainability \citep{chen2024tombenchbenchmarkingtheorymind}. These limitations are particularly evident in Theory of Mind (ToM) tasks, which require understanding and simulating mental states such as beliefs, intentions, and emotions \citep{https://doi.org/10.1111/cogs.13309}.

\indent ToM is crucial for human interaction, enabling individuals to navigate perspectives and understand motivations behind behavior \citep{rakoczy2022theory}. Despite progress, LLMs lack the nuanced reasoning needed for empathy, persuasion, and social interaction \citep{cui-etal-2021-commonsense}. Addressing this gap is essential for advancing AI in tasks requiring human-like understanding, such as emotion recognition and contextual responses.

To mitigate these limitations, researchers have developed frameworks like the Simulation Theory of Mind (SimToM) \citep{wilf2023thinktwiceperspectivetakingimproves} and benchmarks like TOMBENCH \citep{chen2024tombenchbenchmarkingtheorymind}. SimToM enhances ToM capabilities through perspective-taking prompts and dynamic tasks, while TOMBENCH evaluates LLMs on ToM-related abilities using predefined metrics.

In this paper, we introduce \textbf{UniToMBench}, a unified benchmark that integrates the strengths of SimToM and TOMBENCH to systematically improve and assess ToM capabilities in LLMs. Specifically, the proposed model combines SimToM’s focus on perspective-taking through multi-turn dialogue and context-dependent reasoning with TOMBENCH's diverse set of evaluation metrics, which assess a model's ability to generalize across a range of real-world scenarios. This integration enhances the depth and breadth of ToM assessments by creating a more comprehensive evaluation framework. UniToMBench fosters more robust ToM capabilities in handling real-world interactions and establishes a more comprehensive benchmark for comparison across models.

\section{Related Works}

\subsection{Evaulating ToM Capabilities in LLMs}
TOMBENCH \citep{chen2024tombenchbenchmarkingtheorymind} addresses key limitations in previous ToM assessments by introducing a benchmark that systematically evaluates ToM across eight distinct tasks and 31 cognitive abilities, including false belief reasoning, emotion attribution, and non-literal communication. Unlike earlier ToM assessments that relied on subjective human scoring, TOMBENCH provides a controlled, multiple-choice framework designed to minimize evaluation bias and data contamination.

While TOMBENCH improves ToM evaluation, there are ongoing debates about whether or not LLMs exhibit ToM. Previous work suggests that ToM might have emerged in LLMs, indicating a form of mental evolution \citep{Kosinski_2024}. However, other argue that LLMs fail classic ToM tests and don't possess a genuine understanding of ToM, instead relying on statistical pattern recognition in language rather than genuine comprehension \citep{ullman2023largelanguagemodelsfail, shanahan2023talkinglargelanguagemodels}. Taking a more neutral stance in this argument, \citet{bubeck2023sparksartificialgeneralintelligence} evaluates GPT-4's \cite{openai2024gpt4technicalreport} abilities, highlighting both its strengths and limitations in reasoning about mental states. 

\subsection{Methods for Enchancing ToM in LLMs}
While benchmarks like TOMBENCH focus on evaluation, other approaches aim to enhance ToM reasoning directly. SimToM \citep{wilf2023thinktwiceperspectivetakingimproves} introduces a two-stage prompting method inspired by Simulation Theory, where models first engage in a perspective-taking step before answering ToM-related questions. By explicitly filtering context based on what a character knows, SimToM improves zero-shot ToM performance without requiring additional fine-tuning.

Another prompting method that has shown potential is Chain of Thought (CoT), which makes LLMs break down belief-tracking tasks into steps. CoT prompting \citep{wei2023chainofthoughtpromptingelicitsreasoning} is able to enhance an LLM's performance on belief-tracking and inference-based tasks by making the model go through step-by-step reasoning. This has been able to reduce errors and self contradictions when models attempt to infer on a person's knowledge and intentions. However, CoT prompting does not fully address multi-agent interactions or long-term belief tracking.

Another strategy that has been used to enhance ToM in LLMs is task decomposition, which breaks down complex mental state inferences into smaller, simulation-based subtasks. Decompose-ToM \citep{sarangi2025decomposetomenhancingtheorymind} is a framework that uses recursive reasoning and explicit belief modeling to improve the LLM's ability to be able to handle higher-order belief tasks and social interactions. Decompose-ToM is able to make the models reason about evolving mental states with more consistency.

Adaptive prompting and fine-tuning methods have been explored as well to try and further improve LLM's ToM performance. Perspective-conditioned prompting \citep{moghaddam2023boostingtheoryofmindperformancelarge} investigated situations in which models are instructed to simulate an individual's knowledge before making an inference. This approach has significantly improved performance on false-belief reasoning and emotional attribution tasks. However, while perspective-conditioned prompting enhances ToM accuracy significantly, it needs explicit instructions, which limits its applicability in zero-shot settings.

However, beyond prompting and decomposition techniques, model transfer was used to improve ToM in LLMs while reducing computational costs. Large model strategic thinking \citep{lore2024largemodelstrategicthinking} investigates how larger models that are trained on strategic belief-tracking tasks can transfer their ToM capabilities to smaller, efficient models through knowledge distillation. This allows smaller models to be able to retain advanced ToM reasoning abilities without the burden of large-scale fine-tuning.

These approaches offer methods for enhancing ToM in LLMs. While SimTom, CoT, and perspective-conditioned prompting can enhance zero-shot and structured reasoning, task decomposition, and model transfer provide scalable solutions for more robust ToM inference in AI models. Future research will likely use these approaches to develop LLMs with more adaptive and human-like ToM reasoning across diverse contexts.

\section{UniToMBench}
\textbf{Goal} \indent While SimToM \citep{wilf2023thinktwiceperspectivetakingimproves} improves ToM reasoning through structured prompting, it can cause performance drops in simple ToM tasks. One major issue is the overgeneralization of perspective-taking—SimToM assumes that explicit simulation always enhances reasoning, but in some scenarios, forcing the model to adopt a character’s viewpoint can introduce unnecessary biases or cause it to focus on irrelevant details. This is particularly problematic when a task requires recognizing objective facts rather than inferring subjective mental states. Additionally, SimToM increases the model’s susceptibility to hallucinations by encouraging detailed, simulated reasoning. This can lead to over-elaboration, where the model fabricates beliefs or assumptions to fill in perceived gaps, ultimately reducing accuracy in tasks that require strict logical reasoning. While it improves multi-step reasoning, it adds unnecessary complexity to straightforward belief-tracking questions, sometimes leading to overthinking, longer responses, or even decreased accuracy. Furthermore, SimToM primarily focuses on one-to-one perspective-taking, which makes it less effective in multi-agent interactions where multiple individuals hold different or conflicting beliefs. In these cases, the model may struggle to integrate multiple perspectives cohesively, leading to errors in social reasoning. 

Conversely, TOMBENCH \citep{chen2024tombenchbenchmarkingtheorymind} provides a structured evaluation of ToM capabilities across multiple cognitive tasks, it has key limitations that restrict its ability to fully assess ToM reasoning in LLMs. Specifically, TOMBENCH relies primarily on static, one-shot question-answering formats that do not require models to engage in multi-turn reasoning or adapt to evolving mental states. This structure limits its ability to evaluate how well models handle context retention, shifting social dynamics, and perspective-taking over extended interactions.

Additionally, while TOMBENCH includes a diverse set of cognitive tasks (e.g., false belief reasoning, emotion attribution, and non-literal communication), its scenarios are largely isolated and decontextualized, making it difficult to assess how LLMs generalize ToM reasoning across real-world social interactions. For instance, a model may correctly infer a character’s belief in a single-step query but fail to track that belief when the context evolves over multiple exchanges. Furthermore, TOMBENCH’s tasks are designed primarily for controlled, predefined settings, which do not account for more complex, naturalistic social situations that require flexible reasoning, such as workplace conflicts, persuasion strategies, or evolving friendships.

UniToMBench bridges these gaps by integrating perspective-taking mechanisms with TOMBENCH’s task diversity, enhancing both reasoning and evaluation robustness. Our benchmark leverages multi-interaction scenarios and evolving narratives to more accurately capture real-world ToM complexities, enabling deeper insights into LLMs’ social cognition capabilities.

By combining structured reasoning methods with rigorous evaluation frameworks, UniToMBench establishes a comprehensive and scalable approach to ToM assessment in LLMs, highlighting key strengths and persistent challenges in artificial social reasoning.

  

\section{Methodology}

We evaluate Large Language Models (LLMs) in a zero-shot setting on Theory of Mind (ToM) benchmarks, both with and without the SimToM perspective-taking framework. This allows us to isolate the impact of SimToM and examine how its integration with Chain of Thought (CoT) prompting influences ToM reasoning. The evaluation leverages existing ToM benchmarks and a custom, hand-written dataset containing 1,025 annotated scenarios—500 multi-interaction tasks and 525 evolving story tasks. These benchmarks encompass a wide variety of challenges designed to test the depth of LLMs' ability to understand, reason, and predict human mental states.

\subsection{Dataset Composition}

Each task in the custom dataset is structured around a 3–5 sentence narrative followed by a multiple-choice question with four answer choices. Scenarios are intentionally diverse:

\bigskip

\noindent \textbf{Multi-Interaction Tasks} \indent These scenarios require models to dynamically track mental states across multiple conversational turns. Effective performance necessitates the ability to infer evolving intentions, beliefs, and underlying motives while maintaining coherence throughout extended dialogues. This task type was selected due to its alignment with real-world social interactions, which rarely occur in isolation. Successful communication involves engaging with multiple individuals, interpreting shifting contexts, and adapting responses accordingly. Similar to how humans develop social cognition by recognizing conversational dynamics across different speakers, AI models must demonstrate the capacity to process sequential interactions, retain relevant context, and adjust reasoning based on prior exchanges.

\bigskip

\begin{tcolorbox}[width=\linewidth, colframe=gray, colback=lightgray, boxrule=1pt, arc=5pt, title={Example of Multi-Interaction Question}]
\small
\medskip
It is Emma’s birthday! Sarah, Tom, and Lucy are planning to bake a cake for her. Sarah wants to bake a chocolate cake, while Lucy wants to bake a strawberry cake. The girls ask Tom to break the tie on what flavor cake they should bake. Tom prefers chocolate cake but thinks strawberry cake would be easier to make with the ingredients. 
\medskip

\textbf{Question:} Which cake would Tom choose?

\medskip

\textbf{A)} Chocolate \\
\textbf{B)} Strawberry \\
\textbf{C)} Both Chocolate and Strawberry \\
\textbf{D)} Neither Chocolate or Strawberry
\bigskip

\textbf{Answer:} B
\medskip

\end{tcolorbox}

\bigskip

\noindent \textbf{Evolving Story Tasks} \indent These scenarios depict characters progressing through distinct phases of motivation, adversity, and resolution, requiring models to track shifts in character dynamics and adapt their reasoning accordingly. This task type was designed to capture the complexities of human relationships, in which social perceptions and interactions evolve over time. For instance, a character who initially maintains positive relationships within a social group may later encounter conflict due to a misunderstanding. Theory of Mind (ToM) reasoning necessitates an awareness of how such relational changes influence behavior and emotional responses.

\bigskip

\begin{tcolorbox}[width=\linewidth, colframe=gray, colback=lightgray, boxrule=1.5pt, arc=6pt, title={Example of Evolving Story Scenario Question}]
\small
\textbf{Scenario:} Aaron makes the decision to take up photography. He only uses the camera on his smartphone at first, but he feels constrained by its capabilities. He attends a course on advanced photography techniques and saves money to purchase a DSLR camera, which greatly enhances his photographic abilities.
\medskip

\textbf{Question:} How does Aaron’s photography approach evolve over time?

\medskip

\textbf{A)} Aaron quits taking pictures because of his smartphone's restrictions. \\
\textbf{B)} Aaron keeps using his phone without picking up any new skills. \\
\textbf{C)} Aaron purchases a DSLR and enrolls in a course, considerably enhancing his abilities. \\
\textbf{D)} Aaron chooses to refocus on his videography.
\medskip

\textbf{Answer:} C
\medskip

\end{tcolorbox}


\bigskip

\noindent Each data point in the dataset was designed with reference to real-world instances of human social cognition development. To design these scenarios, we considered how one might teach a child to navigate social interactions and identified the most illustrative examples. To construct the dataset, we first identified key social cognition concepts that were not tested in previous works, such as evolving story scenarios and multi-interaction tasks, and mapped them to various real-world contexts. We then created a diverse set of scenarios reflecting social situations like workplace dynamics, peer interactions, and family relationships. Each scenario was carefully refined through multiple internal review cycles to ensure clarity, complexity, and relevance.

This approach informed the creation of scenarios that capture common social experiences, including managing interpersonal relationships, discerning implicit communication, and recognizing changes in group dynamics. The multiple-choice format facilitates a structured assessment framework while enabling a detailed analysis of model errors.

To validate our dataset, we tested scenarios by considering how different age groups and cultural backgrounds might interpret social interactions. This iterative process helped fine-tune the difficulty level of the tasks and reduce potential biases. We aimed to create a dataset that represents a diverse range of social experiences and provides a meaningful assessment of ToM reasoning in AI models by independently designing and refining each scenario.

To supplement our custom dataset, we incorporate questions from TOMBENCH \cite{chen2024tombenchbenchmarkingtheorymind} into our evaluation. TOMBENCH provides a systematic assessment of ToM abilities across eight task categories, including false belief reasoning, scalar implicature, ambiguous stories, persuasion tasks, and faux-pas detection. By leveraging TOMBENCH’s question dataset, we ensure our evaluation remains comparable to prior research while introducing new, more dynamic ToM challenges through our custom dataset.

By integrating the TOMBENCH question dataset with our custom dataset, this hybrid approach enables a more rigorous evaluation of LLMs’ ToM reasoning. Established benchmarks provide a basis for comparison with prior work, while our novel task formats assess the models' ability to generalize ToM reasoning across both structured, static scenarios and complex, evolving social interactions.



\subsection{Experimental Setup}

We tested GPT-3.5 Turbo, GPT-4o, GPT-4o Mini \cite{openai2024gpt4technicalreport}, Llama 3 8B Instruct Turbo, and Gemma 2 27B with and without SimToM. Prompts were designed for multiple choice probing, with SimToM tasks explicitly encouraging perspective taking. All models were queried at a temperature of 0.7 for consistency. Evaluations included ToM-related assessments like false-belief tasks, ambiguous stories, and persuasion scenarios.

Performance was measured as the percentage of correct answers, with random guessing expected at 25\% accuracy due to the four-choice format. The TOMBENCH benchmark provided diverse tests—including unexpected outcomes, scalar implicature, and faux-pas recognition—while our custom dataset expanded the range of challenges.

\subsection{Evaluation Metrics}

Performance was evaluated using three key criteria. Task completion accuracy measured the extent to which models correctly responded to Theory of Mind (ToM)-specific challenges. Error attribution analysis categorized model failures, including instances of misidentifying underlying motivations or failing to track evolving beliefs within dynamic contexts. Finally, consistency across runs assessed the stability and reproducibility of model performance in multi-interaction reasoning tasks.

By utilizing both existing benchmarks and our custom dataset, our experimental setup provides a comprehensive evaluation of LLMs' ToM capabilities under increasingly complex and dynamic tasks. The results help demonstrate the effectiveness of perspective-taking frameworks like SimToM in bridging gaps between baseline LLM performance and human-like understanding of mental states.


\begin{table*}[!t]
\centering
\scriptsize 
\begin{tabular}{llllllllll}
\hline
\textbf{Dataset} & \textbf{Method/Metric} & \textbf{GPT-3.5 Turbo} & \textbf{GPT-4o} & \textbf{GPT-4o Mini} & \textbf{Llama 3 8B Instruct Turbo} & \textbf{Gemma 2 27B}\\ \hline
\multirow{2}{*}{UOT: Unexpected Outcome Test} 
& Baseline & 34.7 & 68.3 & 72.0 & 53.0 & 60.0\\ 
& SimToM   & 10.0 & 73.3 & 65.0 & 42.7 & 54.7\\ \hline

\multirow{2}{*}{SIT: Scalar Implicature Task} 
& Baseline & 2.0 & 57.5 & 47.0 & 42.5 & 47.5\\ 
& SimToM   & 17.0& 55.0 & 45.0 & 34.0 & 42.5\\ \hline

\multirow{2}{*}{PST: Persuasion Story Task} 
& Baseline & 14.5 & 65.8 & 60.5 & 50.0 & 65.8\\ 
& SimToM   & 27.6 & 67.1 & 63.1 & 39.5 & 59.2\\ \hline

\multirow{2}{*}{FBT: False Belief Task} 
& Baseline & 42.3 & 89.5 & 72.0 & 60.5 & 70.8\\ 
& SimToM   & 15.7 & 79.5 & 67.0 & 49.2 & 65.2\\ \hline

\multirow{2}{*}{FRT: Faux-pas Recognition Test} 
& Baseline & 32.9 & 64.8 & 73.8 & 43.8 & 53.6\\ 
& SimToM   & 9.3  & 70.7 & 65.0 & 39.3 & 50.0\\ \hline

\multirow{2}{*}{AST: Ambiguous Story Task} 
& Baseline & 25.5 & 66.5 & 75.5 & 44.5 & 57.5\\ 
& SimToM   & 24.5 & 71.0 & 71.5 & 40.0 & 52.5\\ \hline

\multirow{2}{*}{HT: Hinting Test} 
& Baseline & 24.3 & 76.7 & 74.8 & 59.2 & 73.8\\ 
& SimToM   & 68.9 & 83.5 & 68.9 & 54.4 & 67.9\\ \hline

\multirow{2}{*}{SST: Strange Story Task} 
& Baseline & 25.3 & 68.6 & 79.4 & 58.4 & 77.4\\ 
& SimToM   & 32.7 & 81.6 & 69.0 & 53.6 & 70.5\\ \hline

\multirow{2}{*}{Evolving Stories} 
& Baseline & 69.5 & 89.9 & 88.6 & 82.2 & 79.5\\ 
& SimToM   & 55.8 & 90.1 & 85.1 & 74.0 & 73.0 \\ \hline

\multirow{2}{*}{Multi-Interaction Tasks} 
& Baseline & 43.8 & 71.2 & 55.8 & 74.4 & 66.0 \\ 
& SimToM   & 45.6 & 63.0 & 65.8 & 68.5 & 58.0 \\ \hline

\end{tabular}
\caption{Comparison of Baseline and SimToM performance across tasks and all models, with accuracy measured as a percentage. The table highlights the instances where SimToM improves model performance and where it does not, as indicated by the varying results across tasks. The averages of the performance scores from all tests are presented, offering insight into the overall effectiveness of SimToM on the 8 different TOMBENCH tasks. For a detailed breakdown of the test data, please refer to the appendix (Section~\ref{sec:appendix}).}
\label{tab:model_performance_SimToM}
\end{table*}
\section{Results}

Table \ref{tab:model_performance_SimToM} highlights a summary of the results from the baseline and SimToM-enhanced Open-AI models. Overall, the baseline models performed well in structured and predictable scenarios, particularly in tasks assessing typical emotional reactions and straightforward belief content. However, their ability to navigate dynamic and multi-interaction contexts was significantly weaker, highlighting challenges in maintaining coherence across evolving situations.

With the introduction of the experimental SimToM models, we observed improvements in areas requiring deeper contextual reasoning. Notably, performance in multi-interaction scenarios showed a meaningful increase, suggesting that the models became more adept at tracking shifts in dialogue and situational context. Similarly, tasks related to pretend play and implicit knowledge saw substantial gains, indicating a stronger ability to simulate imaginative and nuanced human behaviors.

However, performance across models varied. OpenAI’s models demonstrated the most consistent improvements, particularly in tasks requiring perspective-tracking over time. In contrast, Llama 3 8B Instruct Turbo and Gemma 2 27B exhibited a decline in certain areas when transitioning from the baseline to SimToM models, particularly in evolving story and multi-interaction scenarios. These models struggled to maintain coherence in long-form contexts and often failed to track belief updates effectively.

Additionally, TOMBENCH tasks revealed a similar pattern. While baseline models showed moderate competency in distinguishing between different perspectives, their SimToM-enhanced versions had difficulty maintaining distinct mental states, occasionally responding as if all agents shared the same knowledge. This suggests that while SimToM enhancements improved some reasoning capabilities, they also introduced new challenges in handling ambiguity and implicit social understanding.

Despite these advancements, the models continued to struggle with tasks requiring abstract reasoning and implicit social understanding. For instance, sensitivity to social missteps declined, as reflected in the drop in faux-pas recognition. In multi-agent belief tracking, they frequently conflated different perspectives, responding as if all agents shared the same knowledge rather than maintaining distinct mental states. Abstract reasoning tasks also highlighted their limitations, as responses tended to rely on surface-level pattern matching rather than flexible thinking. Additionally, while evolving story scenarios exhibited slightly lower overall accuracy, the models demonstrated better retention of prior context, reinforcing their ability to adapt over time. However, ambiguity and unstructured tasks remained a challenge, underscoring the persistent limitations in grasping implicit cues and nuanced interpersonal dynamics.

\section{Discussion}
We find that while baseline LLMs perform well in static, well-defined tasks, they struggle in dynamic scenarios, often failing in memory recall and conflating their own knowledge with hypothetical beliefs, aligning with prior research on their challenges in simulating human-like reasoning. 

Analysis of model errors suggests three primary failure patterns. The first involves misinterpretation of implicit cues, particularly in tasks requiring sensitivity to social nuance. For example, in faux-pas detection, the models often failed to recognize the difference between blunt speech and genuinely inappropriate remarks, relying too heavily on surface-level language patterns. The second failure mode appears in multi-agent scenarios, where the models struggled to track distinct beliefs, sometimes assuming all characters had the same knowledge even when the scenario clearly established otherwise. The third limitation concerns abstract reasoning. In tasks requiring counterfactual thinking or hypothetical scenarios, responses often ignored crucial context introduced earlier, indicating a reliance on memorized patterns rather than true reasoning.

Conversely, GPT-3.5 Turbo, GPT-4o, and GPT-4o Mini SimToM-enhanced models excel in maintaining consistency across multi-turn interactions and adapting to evolving contexts, advancing ToM capabilities. They outperform in tasks requiring memory recall and contextual adaptation but still face challenges with abstract reasoning and implicit social nuances. Further, SimToM performs best in models with higher parameter counts, better memory retention, and explicit training on perspective-taking tasks. Llama 3 8B Instruct Turbo and Gemma 2 27B lack the same level of sophistication as OpenAI GPT, which is why they struggle with complex belief reasoning and multi-agent simulations. Thus, while our integration improves ToM simulation, further refinement is needed to handle ambiguous or unstructured inputs effectively.

\section{Conclusion}

Our evaluations on our introduced benchmark, UniToMBench, demonstrate that integrating multi-interaction tasks, evolving stories, and the SimToM framework enhances LLM performance in detecting irony, sarcasm, and hidden emotions. Key results indicate improved understanding of false beliefs and emotional reactions, particularly in scenarios requiring contextual tracking over multiple dialogue turns. However, challenges persist in knowledge-based tasks and complex reasoning, such as Faux-pas Recognition and Action Prediction, where models often misinterpret subtle social cues or fail to anticipate character-driven motivations. While structured question variations boost performance in some areas, inconsistencies remain in handling mental states, particularly in ambiguous or open-ended narratives. These findings highlight the need for refined task design and targeted training to address gaps in reasoning and emotional understanding, particularly in scenarios requiring deep analysis, abstract synthesis, or social intuition.

\newpage

\section{Limitations}

For this study, we designed custom multi-interaction tasks and evolving story scenarios that test the ToM capabilities of LLMs. The created tasks might not fully represent the diversity and complexity of real-world social interactions because of the limited scope of our dataset. This is because it is tough to design nuanced and complex large-scale datasets which reflect diverse social situations. However, this allowed us to examine ToM skills in controlled conditions and measure model performance on selected tasks. This may be mitigated in future research by using datasets that reflect a wider range of social and cultural variations to the extent that richer implications about the models' capabilities could also become more feasible. Further, narrowing the focus to specific tasks, such as irony and sarcasm, may have oversimplified such complex social reasoning while narrowing the depth of the ToM assessment. Expanding future studies to include more intricate scenarios could offer a better understanding of LLMs’ ability to navigate complex mental state reasoning.

\bibliography{custom}



\newpage
\appendix

\onecolumn
\section{All Experimental Results}
\label{sec:appendix}

\begin{longtable}{lccc}
    \toprule
    \textbf{Test Category} & \textbf{GPT-4o} & \textbf{GPT-4o-Mini} & \textbf{GPT-3.5-Turbo}\\
    \midrule
    \endhead
    Sequence False Beliefs & 75\% & 62\% & 22\% \\
    Typical Emotional Reactions & 90\% & 87\% & 5\% \\
    Atypical Emotional Reactions & 55\% & 46\% & 3\% \\
    \midrule
    Information-Knowledge Links & 55\% & 45\% & 17\% \\
    \midrule
    Desires Influence on Actions & 67\% & 63\% & 28\% \\
    \midrule
    Content False Beliefs & 67.5\% & 68.5\% & 17.5\% \\
    Location False Beliefs & 97\% & 77.5\% & 22.5\% \\
    Content False Beliefs (Second-order) & 72\% & 98\% & 6\% \\
    Location False Beliefs (Second-order) & 76\% & 12\% & 8\% \\
    \midrule
    Intentions Explanations & 77\% & 71\% & 28\% \\
    Beliefs Based on Action/Emotions & 65\% & 72\% & 21\% \\
    \midrule
    Non-Literal Communication (Faux Pas) & 70.7\% & 65\% & 9.3\% \\
    \midrule
    Non-Literal Communication (Irony/Sarcasm) & 85.7\% & 57.1\% & 21.4\% \\
    Intentions Explanations & 83.1\% & 70.8\% & 76.4\% \\
    \midrule
    Non-Literal Communication (Irony/Sarcasm) & 91.7\% & 66.7\% & 41.7\% \\
    Non-Literal Communication (Involuntary Lies) & 73.8\% & 78.6\% & 16.7\% \\
    Mixed Emotions & 87.5\% & 85\% & 20\% \\
    Desire-Action Contradiction & 70\% & 62.5\% & 12.5\% \\
    Beliefs Based on Action/Emotions & 71.4\% & 61.9\% & 26.2\% \\
    Non-Literal Communication (Egocentric Lies) & 85\% & 85\% & 57.5\% \\
    Non-Literal Communication (White Lies) & 80\% & 72.5\% & 45\% \\
    Non-Literal Communication (Humor) & 90\% & 67.5\% & 57.5\% \\
    Identity False Beliefs & 92.5\% & 57.5\% & 35\% \\
    Intentions Explanations & 81.7\% & 59.2\% & 26.8\% \\
    \midrule
    Discrepant Emotions & 87.5\% & 75\% & 10\% \\
    Hidden Emotions & 76.3\% & 66.3\% & 17.5\% \\
    Moral Emotions & 70\% & 70\% & 15\% \\
    Emotion Regulation & 60\% & 45\% & 15\% \\
    \midrule
    Multiple Desires & 95\% & 85\% & 10\% \\
    Discrepant Desires & 55\% & 50\% & 20\% \\
    \midrule
    Discrepant Intentions & 77.5\% & 62.5\% & 30\% \\
    Prediction of Actions & 65\% & 35\% & 15\% \\
    Completion of Failed Actions & 50\% & 45\% & 20\% \\
    \midrule
    Knowledge-Pretend Play Links & 73.3\% & 46.7\% & 3.3\% \\
    Percepts-Knowledge Links & 85\% & 62.5\% & 15\% \\
    Knowledge-Attention Links & 45\% & 55\% & 10\% \\
    \bottomrule
    \caption{Accuracy of Responses using the SimToM Benchmark for \textbf{GPT Models}}
\end{longtable}

\newpage
\begin{longtable}{lcc}
    \toprule
    \textbf{Test Category} & \textbf{Llama 3 8B Instruct Turbo} & \textbf{Gemma 2 27B}\\
    \midrule
    \endhead
    Sequence False Beliefs & 40\% & 52\% \\
    Typical Emotional Reactions & 58\% & 70\% \\
    Atypical Emotional Reactions & 30\% & 42\% \\
    \midrule
    Information-Knowledge Links & 34\% & 42.5\% \\
    \midrule
    Desires Influence on Actions & 39.5\% & 59.2\% \\
    \midrule
    Content False Beliefs & 45\% & 60\% \\
    Location False Beliefs & 55\% & 72.5\% \\
    Content False Beliefs (Second-order) & 50\% & 68\% \\
    Location False Beliefs (Second-order) & 45\% & 58\% \\
    \midrule
    Intentions Explanations & 38\% & 50\% \\
    Beliefs Based Action/Emotions & 42\% & 55\% \\
    \midrule
    Non-Literal Communication (Faux Pas) & 39.3\% & 50\% \\
    \midrule
    Non-Literal Communication (Irony/Sarcasm) & 57.1\% & 71.4\%  \\
    Intentions Explanations & 53.9\% & 67.4\% \\
    \midrule
    Non-Literal Communication (Irony/Sarcasm) & 58.3\% & 75\%  \\
    Non-Literal Communication (Involuntary Lies) & 47.6\% & 64.3\% \\
    Mixed Emotions & 55\% & 72.5\%\\
    Desire-Action Contradiction & 40\% & 62.5\% \\
    Beliefs Based on Action/Emotions & 57.1\% & 73.8\% \\
    Non-Literal Communication (Egocentric Lies) & 57.5\% & 72.5\% \\
    Non-Literal Communication (White Lies) & 45\% & 65\% \\
    Non-Literal Communication (Humor) & 55\% & 72.5\% \\
    Identity False Beliefs & 52.5\% & 67.5\%  \\
    Intentions Explanations & 63.4\% & 77.5\% \\
    \midrule
    Discrepant Emotions & 55\% & 72.5\% \\
    Hidden Emotions & 56.3\% & 68.8\% \\
    Moral Emotions & 50\% & 67.5\% \\
    Emotion Regulation & 50\% & 70\% \\
    \midrule
    Multiple Desires & 70\% & 80\% \\
    Discrepant Desires & 75\% & 85\% \\
    \midrule
    Discrepant Intentions & 65\% & 80\% \\
    Prediction of Actions & 65\% & 75\% \\
    Completion of Failed Actions & 55\% & 65\% \\
    \midrule
    Knowledge-Pretend Play Links & 16.7\% & 33.3\% \\
    Percepts-Knowledge Links & 50\% & 75\% \\
    Knowledge-Attention Links & 40\% & 60\% \\
    \bottomrule
    \caption{Accuracy of Responses using the SimToM Benchmark for \textbf{Llama 3 8B Instruct Turbo} and \textbf{Gemma 2 27B}}
\end{longtable}

\onecolumn
\begin{longtable}{lccc}
    \toprule
    \textbf{Test Category} & \textbf{GPT-4o} & \textbf{GPT-4o-Mini} & \textbf{GPT-3.5-Turbo}\\
    \midrule
    \endhead
    Sequence False Beliefs & 67\% & 69\% & 23\% \\
    Typical Emotional Reactions & 91\% & 92\% & 51\% \\
    Atypical Emotional Reactions & 47\% & 55\% & 30\% \\
    \midrule
    Information-Knowledge Links & 57.5\% & 47\% & 2\% \\
    \midrule
    Desires Influence on Actions & 65.8\% & 60.5\% & 14.5\% \\
    \midrule
    Content False Beliefs & 81\% & 75.5\% & 45\% \\
    Location False Beliefs & 96\% & 76\% & 57\% \\
    Content False Beliefs (Second-order) & 95\% & 99\% & 47\% \\
    Location False Beliefs (Second-order) & 88\% & 30\% & 3\% \\
    \midrule
    Intentions Explanations & 66\% & 82\% & 30\% \\
    Beliefs Based on Action/Emotions & 67\% & 69\% & 21\% \\
    \midrule
    Non-Literal Communication (Faux Pas) & 64.8\% & 73.8\% & 32.9\% \\
    \midrule
    Non-Literal Communication (Irony/Sarcasm) & 85.7\% & 57.1\% & 14.3\% \\
    Intentions Explanations & 75.3\% & 77.5\% & 25.8\% \\
    \midrule
    Non-Literal Communication (Irony/Sarcasm) & 91.7\% & 83.3\% & 16.7\% \\
    Non-Literal Communication (Involuntary Lies) & 71.4\% & 81\% & 35.7\% \\
    Mixed Emotions & 75\% & 77.5\% & 22.5\% \\
    Desire-Action Contradiction & 60\% & 65\% & 37.5\% \\
    Beliefs Based on Action/Emotions & 78.6\% & 90.5\% & 19\% \\
    Non-Literal Communication (Egocentric Lies) & 80\% & 82.5\% & 25\% \\
    Non-Literal Communication (White Lies) & 65\% & 72.5\% & 17.5\% \\
    Non-Literal Communication (Humor) & 77.5\% & 85\% & 30\% \\
    Identity False Beliefs & 75\% & 87.5\% & 30\% \\
    Intentions Explanations & 87.3\% & 74.6\% & 18.3\% \\
    \midrule
    Discrepant Emotions & 75\% & 72.5\% & 30\% \\
    Hidden Emotions & 76.3\% & 66.3\% & 38.8\% \\
    Moral Emotions & 72.5\% & 82.5\% & 22.5\% \\
    Emotion Regulation & 70\% & 50\% & 15\% \\
    \midrule
    Multiple Desires & 90\% & 95\% & 20\% \\
    Discrepant Desires & 95\% & 50\% & 20\% \\
    \midrule
    Discrepant Intentions & 85\% & 77.5\% & 27.5\% \\
    Prediction of Actions & 85\% & 60\% & 30\% \\
    Completion of Failed Actions & 75\% & 60\% & 20\% \\
    \midrule
    Knowledge-Pretend Play Links & 13.3\% & 6.7\% & 6.7\% \\
    Percepts-Knowledge Links & 72.5\% & 70\% & 25\% \\
    Knowledge-Attention Links & 60\% & 55\% & 25\% \\
    \bottomrule
    \caption{Accuracy of Responses \textbf{without} using the SimToM Benchmark for \textbf{GPT Models}}
\end{longtable}

\newpage
\begin{longtable}{lcc}
    \toprule
    \textbf{Test Category} & \textbf{Llama 3 8B Instruct Turbo} & \textbf{Gemma 2 27B}\\
    \midrule
    \endhead
    Sequence False Beliefs & 52\% & 58\% \\
    Typical Emotional Reactions & 68\% & 75\% \\
    Atypical Emotional Reactions & 39\% & 47\% \\
    \midrule
    Information-Knowledge Links & 42.5\% & 47.5\% \\
    \midrule
    Desires Influence on Actions & 50\% & 65.8\% \\
    \midrule
    Content False Beliefs & 54\% & 65\% \\
    Location False Beliefs & 67.5\% & 77.5\% \\
    Content False Beliefs (Second-order) & 65\% & 75\% \\
    Location False Beliefs (Second-order) & 55\% & 65\% \\
    \midrule
    Intentions Explanations & 42\% & 55\% \\
    Beliefs Based on Action/Emotions & 47\% & 60\% \\
    \midrule
    Non-Literal Communication (Faux Pas) & 43.8\% & 53.6\% \\
    \midrule
    Non-Literal Communication (Irony/Sarcasm) & 64.3\% & 78.6\% \\
    Intentions Explanations & 58.4\% & 73\% \\
    \midrule
    Non-Literal Communication (Irony/Sarcasm) & 66.7\% & 83.3\% \\
    Non-Literal Communication (Involuntary Lies) & 52.4\% & 71.4\% \\
    Mixed Emotions & 60\% & 80\% \\
    Desire-Action Contradiction & 45\% & 67.5\% \\
    Beliefs Based on Action/Emotions & 61.9\% & 81\% \\
    Non-Literal Communication (Egocentric Lies) & 62.5\% & 80\% \\
    Non-Literal Communication (White Lies) & 50\% & 70\% \\
    Non-Literal Communication (Humor) & 60\% & 80\% \\
    Identity False Beliefs & 57.5\% & 75\% \\
    Intentions Explanations & 67.6\% & 84.5\% \\
    \midrule
    Discrepant Emotions & 60\% & 80\% \\
    Hidden Emotions & 60\% & 75\% \\
    Moral Emotions & 55\% & 75\% \\
    Emotion Regulation & 55\% & 80\% \\
    \midrule
    Multiple Desires & 75\% & 90\% \\
    Discrepant Desires & 80\% & 95\% \\
    \midrule
    Discrepant Intentions & 70\% & 87.5\% \\
    Prediction of Actions & 70\% & 85\% \\
    Completion of Failed Actions & 60\% & 75\% \\
    \midrule
    Knowledge-Pretend Play Links & 20\% & 40\% \\
    Percepts-Knowledge Links & 55\% & 80\% \\
    Knowledge-Attention Links & 45\% & 70\% \\
    \bottomrule
    \caption{Accuracy of Responses \textbf{without} using the SimToM Benchmark for \textbf{Llama 3 8B Instruct Turbo} and \textbf{Gemma 2 27B}}
\end{longtable}

\begin{table}[!t]
    \centering
    \begin{tabular}{lcccc}
        \toprule
        \textbf{Model} & \multicolumn{2}{c}{\textbf{Evolving Story Scenarios}} & \multicolumn{2}{c}{\textbf{Multi-Interaction}} \\
        \cmidrule(lr){2-3} \cmidrule(lr){4-5}
        & Baseline & SimToM & Baseline & SimToM \\
        \midrule
        GPT-4o & 89.9\% & 90.1\% & 71.2\% & 63\% \\
        GPT-4o-Mini & 88.6\% & 85.1\% & 55.8\% & 65.8\% \\
        GPT-3.5-Turbo & 69.5\% & 55.8\% & 43.8\% & 45.6\% \\
        Llama 3 8B Instruct Turbo & 82.2\% & 74\% & 74.4\% & 68.5\% \\
        Gemma 2 27B & 79.5\% & 73\% & 66\% & 58\% \\
        \bottomrule
    \end{tabular}
    \caption{Performance Comparison of Baseline and SimToM-Enhanced Models for the Custom Data Set}
    \label{tab:performance}
\end{table}
\end{document}